\pdfoutput=1
\documentclass[11pt,a4paper]{article}
\usepackage{eacl2023}
\usepackage[american]{babel}
\usepackage{times}
\usepackage[T1]{fontenc}
\usepackage{amssymb}
\usepackage{amsmath}
\usepackage{microtype}
\usepackage{url}
\usepackage{multirow} 
\usepackage{array}
\usepackage{microtype}
\usepackage{enumitem}

\usepackage[utf8]{inputenc}

\RequirePackage{xcolor}
\definecolor{lightgray}{gray}{0.80}
\definecolor{mediumgray}{gray}{0.60}
\definecolor{darkgray}{gray}{0.40}

\usepackage[pdftex]{graphicx}
\usepackage{booktabs}
\DeclareGraphicsExtensions{.pdf,.ai,.jpg,.png}
\setkeys{Gin}{pagebox=artbox}
\graphicspath{{./eacl23-conclusion-based-counter-argument-generation-figures/}}

\newcommand{\hwfigure}[3][t]{
	\begin{figure*}[#1]
		\centering
		\includegraphics[scale=1.0]{#2}
		\caption{#3}\label{#2}%
	\end{figure*}
}

\RequirePackage{type1cm}
\RequirePackage{color}
\RequirePackage{soul}
\setstcolor{blue}
\definecolor{violet}{rgb}{0.5,0.0,0.5}
\newsavebox\bscombox
\newcommand{\bscom}[3][]{%
	\sbox{\bscombox}{\fontsize{8}{9}\selectfont#1#2#3}
	\noindent
	\st{#2}{\selectfont
		\color{blue}#3\ifx\\#1\\\else{\fontsize{8}{9}\selectfont\color{violet}[#1]}\fi
	}
}

\begin{document}

%
%

\title{Conclusion-based Counter-Argument Generation}

\newcommand{\dlr}{\textsuperscript{$\ddagger$}}
\newcommand{\pb}{\textsuperscript{$\dagger$}}

\author{%
	Milad Alshomary \pb
	\qquad Henning Wachsmuth \pb \\[1.5ex] 
	\pb Leibniz University Hannover, Institute of Artificial Intelligence \\[1.5ex]
	{\tt m.alshomary@ai.uni-hannover.de}\\
}

\maketitle

\begin{abstract}
In real-world debates, the most common way to counter an argument is to reason against its main point, that is, its conclusion. Existing work on the automatic generation of natural language counter-arguments does not address the relation to the conclusion, possibly because many arguments leave their conclusion implicit. In this paper, we hypothesize that the key to effective counter-argument generation is to explicitly model the argument's conclusion and to ensure that the stance of the generated counter is opposite to that conclusion. In particular, we propose a multitask approach that jointly learns to generate both the conclusion and the counter of an input argument. The approach employs a stance-based ranking component that selects the counter from a diverse set of generated candidates whose stance best opposes the generated conclusion. In both automatic and manual evaluation, we provide evidence that our approach generates more relevant and stance-adhering counters than strong baselines.
\end{abstract}

\section{Introduction}
\label{sec:introduction}

Given an argument, a valid counter-argument should be relevant to the topic discussed by the argument while opposing to its conclusion's stance. Countering the opponent's arguments in a debate effectively is key to winning the debate \cite{zhang:2016}. While some counter-arguments attack an argument's premises or their connection to the conclusion, the most common way is to directly rebut the argument's conclusion \cite{walton:2009}.

The automatic countering of natural language arguments is one of the most challenging tasks in the area of computational argumentation. Previous work has addressed the task through retrieval \cite{wachsmuth:2018a,orbach:2020} or generation-based approaches \cite{hua:2018, hidey:2019}. By concept, the former requires the presence of suitable counter-arguments in a predefined collection, limiting its flexibility. Existing generation-based approaches, on the other hand, either consider a single claim as input or do not model the relation between premise and conclusion in the input argument. In daily-life debates, however, people often do not explicitly state their argument's main point (i.e., its conclusion), since it is often clear from the context \cite{habernal:2015}. This makes it challenging for computational models to generate a proper counter. Table 1 shows an example of an argument with the conclusion ``Purchasing meat is encouraging animal abuse''. The author states that meat production would often lead to animal abuse. However, this statement is never linked explicitly to the conclusion. Such a link may be easy to infer for humans, but it is challenging for machines.

\begin{table}[t!]%
	\centering%
	\small
	\renewcommand{\arraystretch}{1.0}
	\setlength{\tabcolsep}{2.5pt}%
	\begin{tabular}{p{0.95\columnwidth}}
		\toprule
		{\bf Conclusion (title)}: Purchasing meat encourages animal abuse. \\
		\midrule
		{\bf Premises (post):} All meat, to my knowledge, is obtained by raising animals in cramped quarters and slaughtering them as soon as they are fully grown. The only exception i can think of is perhaps when you go into the woods and hunt food for yourself in which case the animal has lived an undisturbed life and is put down by humane means compared to how it happens in nature. However, this is, of course, time intensive, requires skill, expensive, and thus is of course not how the vast majority of meat is obtained.\\
		\bottomrule
	\end{tabular} 
	\caption{An example argument (conclusion + premises) taken from {\em Reddit ChangeMyView}, showing how the conclusion is mentioned implicitly only in the body.}
	\label{table-intro-example}
\end{table}

State-of-the-art transformer-based language models excel in many downstream text generation tasks, such as summarization and machine translation \cite{vaswani:2017}. While they have been applied successfully for reconstructing implicit argument components such as conclusions \cite{gurcke:2021,syed:2021}, even they still fall short on more complex tasks, such as counter-argument generation \cite{hua:2019}. 

In this paper, we study how to enable transformer-based language models to generate an effective counter-argument to a given argument. We observe that the performance of these models in generating relevant counters with correct stance deteriorates particularly when the input argument does not mention its conclusion. Hence, we hypothesize that explicitly modeling the argument's conclusion and its stance will lead to more adequate counter-arguments. For this purpose, we propose a multitask generation approach with a stance-based ranking component. Our approach jointly models the two tasks of conclusion generation and counter-argument generation, and it ensures stance correctness through a stance-based ranking component.

Given a training dataset, where we have access to both the premises of arguments and their corresponding conclusions and counters, we explore two variations of the approach: The first shares the transformer's encoder and decoder between the two tasks, and we learn to generate both the conclusion and the counter as one sequence (separated with a special token). By contrast, the second variation is composed of one shared encoder along with two decoders, one to generate the conclusion and the other to generate the counter-argument. Although we expect the trained models to often capture the stance relation between the argument and its counter, we reinforce opposite stance through a stance-based ranking component at inference time. This component samples different counter-arguments and ranks them based on their stance score towards the corresponding generated conclusion. 

To evaluate both approach variations, we use the ChangeMyView dataset of \cite{jo:2020} that consists of discussions where someone posts a view and others write comments opposing to this view. In line with related work \cite{alshomary:2021}, we use a post's title as the conclusion, its body text as the premises, and each comment as a counter. To classify stance as part of our ranking component, we fine-tune RoBERTa \cite{liu:2019} on a dataset of pairs of claim and counter-claim collected from the {\em Kialo.org} debate platform. We compare the approach variations against two baselines; one that learns to generate the conclusion and the counter-argument independently in a pipeline model and one that employs a sequence-to-sequence model but does not learn actively to represent the conclusion. The results demonstrate the deficiency of standard transformer-based models, particularly when the conclusion is not mentioned explicitly, highlighting the importance of conclusions in counter-argument generation. In most cases, our variation with shared encoder and decoder produces the best counter-arguments in terms of relevance and stance correctness. 

We summarize our contributions as follows:\footnote{Code is attached in the materials, and will be made publicly available upon acceptance.}
\begin{itemize}
	\item We study how to generate effective counter-arguments even if the attacked argument's conclusion is implicit.
	\item We present two multitask transformer-based counter-argument approaches, tuned to opposing to the argument's conclusion.
	\item We reveal the impact of modeling an argument's conclusion and counter-argument jointly in the given task.
\end{itemize}

\section{Related Work}

Argument generation is one of the main branches of computational argumentation, studying the synthesis of arguments in natural language texts. This field includes a host of tasks like the generation of argument conclusions \cite{alshomary:2020, syed:2021}, implicit premises \cite{chakrabarty:2021}, controlled claims \cite{schiller:2020, alshomary2021b}, as well as the generation of counter-arguments \cite{hua:2018, alshomary:2021}. Our work studies the task of counter-argument synthesis.

The task of counter-argument synthesis has been addressed through either retrieval or generation-based approaches. An example of the former is the work of \citet{orbach:2020} whose approach tries to retrieve relevant counters for a given argument from a collection of documents. \citet{wachsmuth:2018a} utilized topic knowledge to retrieve the best counter for a given argument. 

On the other hand, generation-based approaches aim to construct counter-arguments from scratch. For example, both \citet{bilu:2015} and \citet{hidey:2019} worked on the task of counter-claim generation. The former developed a set of rules and classifiers to negate claims, while the latter used neural methods to learn from data. \citet{alshomary:2021} proposed an approach to generate counter-arguments by automatically identifying weak points in the input argument given the conclusion and attacking them. Moreover, \citet{hua:2018,hua:2019} proposed an approach for generating long texts and applied it to the counter-argument generation task. Their approach relies on a retrieval component that acquires relevant key phrases for an input argument to be used to guide the generation of counter-arguments. While the size of the given argument collection limits retrieval-based approaches, the generation-based approaches either rely on the conclusion being given in the input or don't distinguish the different components in the input argumentative text. Our proposed approach is generation-based, where we study the conclusion's role in the counter-argument generation task.

Argument conclusion is the main point an argument argues towards/against, which is important for understanding the argument. In daily life argumentation, conclusions often are left implicit \citet{alshomary:2020}. While it is easy for humans to infer the main point of an argument, it remains a challenging task for machines. Hence, several works have addressed the task of conclusion inference. \citet{alshomary:2020} reconstructed implicit claim targets from argument premises using triplet neural networks. \citet{syed:2021} studied the effectiveness of several transformer-based models on the conclusion generation and evaluated the informativeness criteria of conclusions. \citet{gurcke:2021} utilized conclusion generation to study argument quality. Our proposed approach also generates conclusions for a given argument as the first step in order to generate reliable counters.

\section{Approach}

\hwfigure{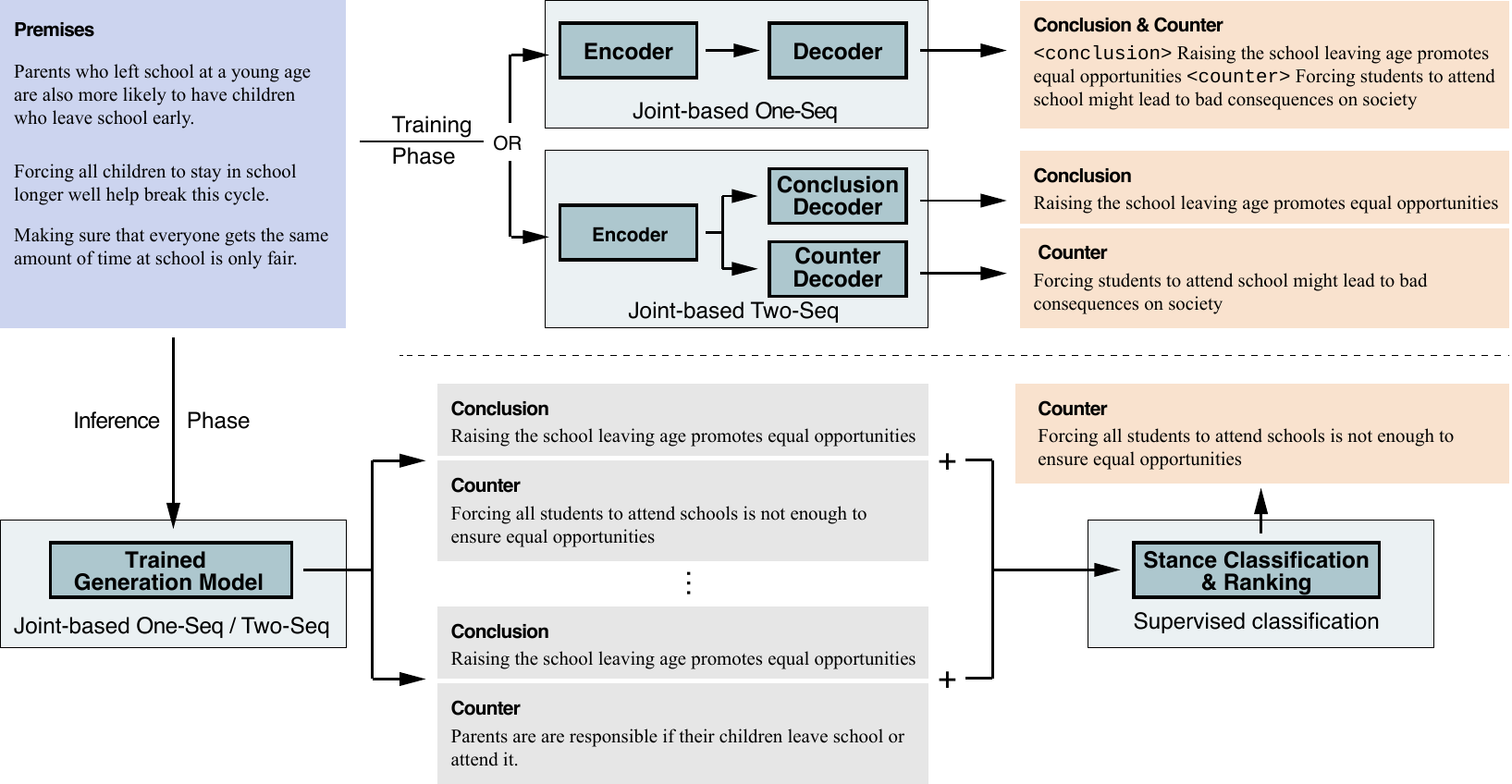}{Proposed approach to counter-argument generation. In the \emph{training phase}, we learn to jointly generate the conclusion and counter either as one sequence (\textit{Joint-based One-seq}) or as two separated sequences (\textit{Joint-based Two-decoders}). In the \emph{inference phase}, we classify and rank a diverse set of counters with respect to their stance towards the corresponding conclusion.}

As discussed above, the conclusions of arguments are important for understanding them properly. However, they are often left implicit, making understanding hard for machines. Our goal is to study how the absence of conclusions affects the performance of transformer-based counter-argument generation models. To alleviate this problem, we propose an approach that jointly learns to generate both the conclusion and the counter of an argument. At inference time, it utilizes a stance-based ranking component to select the most contrastive candidate counter in each case. We detail the generation and ranking in the following.

\subsection{Joint Generation of Conclusions and Counters}
Text generation is usually modeled as a sequence-to-sequence generation task and is widely addressed through transformer-based encoder-decoder models \cite{vaswani:2017}. Since we aim to learn two generation tasks (conclusion and counter), one could think of either sharing the full model between the two tasks or only the encoder part. Hence, as illustrated in Figure 1, we experiment with both options to realize our approach:

\paragraph{Fully-shared Encoder and Decoder}

In the first model, we maintain the same transformer-based encoder-decoder architecture and train it to generate output sequences containing both the conclusion and the counter. Hence, the model learns to perform the two tasks simultaneously. Particularly, the input to the model is one sequence representing an argument's premises, and the output is a single sequence composed of the ground-truth conclusion and counter-argument separated by special tokens, \texttt{<conclusion>} and \texttt{<counter>}. The model encodes premises and decodes first the conclusion and then the counter in one sequence. We train the model to optimize the following loss function:
$$
L(\theta) = - \sum_{1}^{n} \log p(y_i|X, y_{<i};\theta)
$$

Where  $X$ is the input sequence representing premises, $y$ is the sequence composing the conclusion and counter, and $\theta$ is the model's parameters. We call this model {\em Joint One-seq} later in our experiments. 

At inference time, we utilize a mechanism to generate a diverse set of $n$ candidate conclusions and their counter-arguments, which are later passed to our stance-based ranking component to select the best counter. The diverse generation is as follows. We first extract a set of $m$ Wikipedia concepts from the input premises using the approach of \citet{dor:2018}. Then, during decoding, we use these concepts to prompt our trained model by masking all logits except the ones matching the prompt tokens, resulting in conclusions addressing different aspects of the premises followed by their corresponding counters. Moreover, to ensure candidate diversity, we enable nucleus sampling \cite{holtzman:2019}, where at each step, we randomly select one of the top $k$ tokens with an accumulated probability of more than $p$.

\paragraph{Shared Encoder with two Decoders}

Similarly, the second model starts with an argument's premises as input. However, it then decodes two independent sequences representing the conclusion and the counter-argument as output. First, the input premises are passed through a shared encoder, and then two decoders are used to learn to generate the counter and the conclusion. During training, we optimize the following multi-task loss function, which is a weighted average of the two language modeling losses of the two decoders:%
\footnote{The best weights, $\alpha_{a}$ and $\alpha_{b}$ are determined experimentally during validation.}
\begin{alignat}{2}
&L(\theta_e, \theta_a, \theta_b) \;=\; & \alpha_{a} \cdot \sum_{1}^{n} \log p(y^{a}_{i}|X, y^{a}_{<i};\theta_e;\theta_a) \\ 
& &+\; \alpha_{b} \cdot \sum_{1}^{m} \log p(y^{b}_{i}|X, y^{b}_{<i};\theta_e;\theta_b)
\end{alignat}

Here, $y^{a}$ and $y^{b}$ are the conclusion and counter sequences. $\theta_e$ and $\theta_a$, and $\theta_b$ are the weight parameters of the encoder, the conclusion, and counter decoders, respectively.

The distinction between this model and the previous one is the shared layers between the two tasks. In the previous model, both the encoder and decoder layers are shared between the two tasks, while, here, only the encoder's layers are shared, keeping a dedicated decoder for each of the two tasks. We refer to this model as {\em Joint Two-seq} below.

We aim to generate a diverse set of candidate counters similar to the above model. However, we noticed that counters rarely start by referring to entities or similar concepts, and prompting the model with concepts might lead to generating irrelevant texts. Hence, we generate one conclusion for this model, but a set of candidate counters by only enabling the nucleus sampling during decoding.%
\footnote{We tested the performance of the model empirically and noticed that these prompted counters of low quality.}

\subsection{Ranking Component}

Give a set of $n$ generated candidate counters, we rank them based on their stance contrastiveness towards the corresponding generated conclusion and select the top-ranked as our final output. In particular, we trained a transformer-based stance classifier on pairs of claim and counter-claim acquired from the {\em kialo.com} platform to be used to predict whether the pair have a {\em pro} or {\em con} stance. Experimental details are provided in the next section. To guarantee stance coherence of the selected counter, we compute the stance-based scores on the sentence level to ensure all sentences have some degree of contrastiveness towards the conclusion. In particular, given a pair of a conclusion and the corresponding counter, we first split the counter into a set of sentences. For each sentence $s_i$, we apply our trained classifier to compute the stance $label$ towards the conclusion $c$  and its probability $pr_{label}$. We then translate this into a stance contrastiveness score as follows:
\[cont(s_i, c)=\begin{cases}
\text{$pr_{con}$}, & \text{if $label=con$}\\
\text{$-pr_{pro}$}, & \text{if $label=pro$} 
\end{cases}
\]

The final score of a counter is averaged across its sentences, ranging from -1 to 1. The counters are then ranked accordingly, selecting the top one.
\section{Experiments}

This section describes the experiments carried out to investigate the conclusion's importance in counter-argument generation.

\subsection{Data}

We evaluate our approach on the ChangeMyView (CMV) dataset of \citet{jo:2020}. On the CMV platform, users publish their opinions on controversial topics as posts consisting of a title summarizing the main point and a body representing the reasoning behind it. In turn, others comment on these posts trying to convince the authors to change their mind. We follow \citet{alshomary:2021} by assuming the following mapping: The title of a post represents an argument's {\em conclusion} and its body is the {\em premises}, while each comment is a {\em counter-argument}. To ensure our models are trained on high quality counters, we select for each post the comment with highest argumentative quality score predicted by the model proposed by \citet{gretz:2020b}.

To study counter-argument generation for settings where the conclusion is not mentioned explicitly, we use only the post's body as input, and the title as training output to learn to generate the conclusion. Since users might also restate their post's main point (the conclusion) inside their post, this allows us to study and evaluate the correlation between a model's effectiveness in generating good counter-arguments and the level of implicitness of the conclusion in the input.

The stance-based ranking component relies on a classifier that assesses the stance polarity between two statements. To train such a classifier, we use dataset of \citet{syed:2021}, which is based on the {\em Kialo.org} platform, where claims on controversial topics contributed by humans are organized in a hierarchical structure with supporting and opposing relations. We transformed the data into pairs of claims labeled as \emph{pro} or \emph{con}, and we split it by debates into 95.6k instances for training, 7.7k for validation, and 22.4k for testing.

\subsection{Models}


\paragraph{Approach}

For generation, we used BART as our base model \cite{lewis:2020}, and fine-tuned it starting from the \emph{BART-large} checkpoint. We trained for three epochs using a learning rate of $5^e-5$ and a batch size of $8$. We then selected the checkpoint with the lowest error on the validation set. To find the best parameters $\alpha_{a}$ and $\alpha_{b}$ for the {\em Joint Two-seq} model, we explored pairs of values between 0.1 and 1.0 on a sample of the training set, and took the pair that led to the lowest validation loss: $\alpha_{a}=0.7$ and $\alpha_{b}=0.3$.


To obtain a diverse set of candidate counters for ranking, we used nucleus sampling \cite{holtzman:2019} with $p=0.95$ and $top\_k=50$. For the \textit{Joint One-seq} model, we obtained relevant Wikipedia concepts from the input premises using Project Debater's API\footnote{https://github.com/IBM/debater-eap-tutorial} that we used to prompt the output sequence (conclusion and counter-argument) to encourage diversity. As for the stance classifier, we fine-tuned \textit{roberta-large} on the Kialo pairs for three epochs with learning rate $2e^{-5}$ and batch size $64$. The trained classifier achieved an F$_1$-score of 0.81 on the test split. To test its performance on the ChangeMyView data, we took a sample of 2k instances with pro pairs (an argument and its conclusion) and con pairs (conclusion and counter). The trained classifier resulted in an F$_1$-score of 0.70.

\paragraph{Baselines} 

To study the effectiveness of transformer-based models when the conclusion is not explicitly stated, we compare against four BART-based models, all trained on the conclusion and premises as input and the counter-argument as output, but treated differently in the inference time. 

In particular, the first baseline  ({\em BART-based w/o Conclusion}) relies only on the premises at inference time. To account for the missing conclusion, the second ({\em Pipeline-based}) generates a conclusion using another BART-based conclusion generation model trained independently on the training split of the CMV dataset. This can be seen a pipeline alternative to our approach, since conclusions and counters are learned independently. We also evaluate a variation of this pipeline approach that chooses the best counter among a diverse set of candidates using our ranking component ({\em Pipeline-based w/ Stance}). Finally, the fourth model is an oracle that knows the ground-truth conclusion in addition to the premises ({\em BART-based w/ Conclusion}).

Additionally, we compare our approach with the argument undermining approach of \citet{alshomary:2021} in which the argument's weak points are first identified subject to its conclusion. Then a counter is generated to attack the weakest point(s). We obtained the trained model from the authors and used it to generate counter-arguments corresponding to the top three weak points (similar to their experiments).

\subsection{Automatic Evaluation}

In the following we introduce the automatic evaluation measures used in our experiments. We then present the evaluation results of our approaches, as well as a detailed analysis of their effectiveness with respect to argument length (measured by number of tokens) and conclusion implicitness.

\begin{table}[t!]%
	\centering%
	\small
	\renewcommand{\arraystretch}{1.}
	\setlength{\tabcolsep}{1.25pt}%
	\begin{tabular}{lrr@{\,\,}rr}
		\toprule
		\bf Approach 				 		 & \bf BLEU & \bf Be.F$_1$ 	& \bf Stance 	& \bf Contr. \\
		\midrule
		{BART-based w/o Conclusion}  		 & 0.149 		& 0.138 		& 0.814 		& 0.447\\
		{Pipeline-based}					 & 0.148 		& 0.142 		& 0.816 		& 0.437 \\
		{Pipeline-based w/ Stance}			 & 0.141 		& 0.142 		& 0.852 		& 0.615 \\
		\addlinespace
		{Joint One-seq}				 & 0.143		& *\bf 0.159 	& 0.850			& *0.480 \\
		{Joint One-seq w/ Stance}		 & 0.140        & *0.147        & \bf 0.889     & *\bf 0.661 \\
		{Joint Two-decoders} 			 & *0.154 		& *0.148 		& 0.798 		& 0.423 \\
		{Joint Two-decoders w/ Stance} & \bf *0.164 	& *0.153 		& 0.825 		& *0.652 \\
		\midrule
		{BART-based w/ Conclusion}			 & 0.175		& 0.160			& 0.773			& 0.584 \\
		Argument Undermining				 & 0.072        & 0.090			& 0.805			& 0.664  \\
		
		\bottomrule
	\end{tabular}%
	\caption{Automatic evaluation of our two models, with and without \emph{stance} ranking, compared to baselines, in terms of the similarity of the generated and the ground-truth counters (BLEU and BERT F$_1$-score) and of the counter's correct (opposing) stance. Stance is computed once using Project Debater's API (\emph{Stance}) and once with our stance classifier (\emph{Contrastiveness}). Results highlighted with * are significantly better than the {BART-based w/o Conclusion} with confidence level of 95\%.}
	\label{data-argument-examples-table}%
\end{table}

\begin{figure}
	\centering
	\includegraphics[scale=0.9]{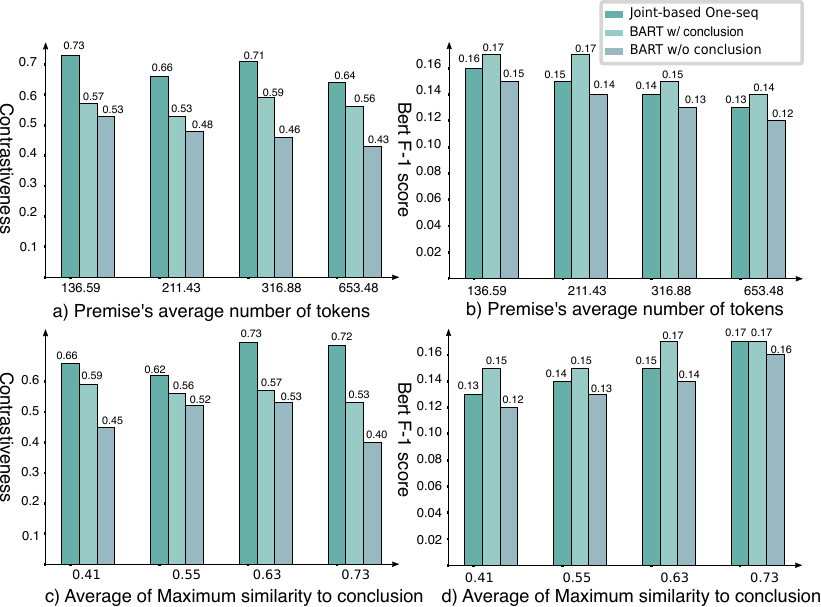}
	\caption{Contrastiveness and BERT F$1$-scores of our approach against the baseline subject to different levels of argument complexity (measure by number of tokens) and conclusion implicitness (measures by maximum similarity of the ground-truth conclusion to the premises).}
	\label{analysis-graphs}
\end{figure}

\paragraph{Evaluation Measures} 

To approximate the similarity of generated and ground-truth counters, we compute BLEU and BERT F$_1$-score\footnote{For each instance, we compare against all ground-truth counters and take the maximum score achieved}. In addition, we measure the stance correctness of the generated counter with respect to the ground-truth conclusion in two ways: First, a {\em contrastiveness} score is computed using the stance classifier trained for our ranking component. It represents the average likelihood of classifying the counter and the corresponding ground-truth conclusion as \emph{con} across the evaluation dataset. Second, a target-based {\em stance} score measures the stance of both the conclusion and the counter towards the conclusion target. Given the valdiation set, we extract the target of each conclusion for this purpose as proposed by \citet{alshomary:2020} and then use Project Debater's API%
\footnote{\url{https://early-access-program.debater.res.ibm.com/}}
to classify the conclusion's stance and the generated counter's stance towards the extracted target. The final measure is the absolute difference between the counter and conclusion scores, averaged across the evaluation dataset.

\paragraph{Results} 

Table 1 shows the evaluation results. All approaches are close in BLEU and BERT F$1$-score, with small but significant advantages for our models. We observe that the absence of explicit mention of the conclusion in the input ({\em Bart-based w/o Conclusion}) worsens the results across all measures but the Stance score, and vice versa when introducing the conclusion ({\em Bart-based w/ Conclusion}). This clearly indicates the importance of the conclusion in the process of counter-argument generation. 

When the conclusion is not mentioned explicitly but has to be inferred, we can see that both our generation models which jointly generate conclusions and counters, outperform the baselines in terms of correct stance. As expected, adding the ranking component to our approaches and the pipeline baseline consistently boosts the correctness, the best being \emph{Joint One-seq w/ Stance} with stance score 0.889 and contrastiveness score 0.661.

Although the Argument Undermining approach of \citet{alshomary:2021} requires an explicit mention of the conclusion to rank premises according to their attackability, its effectiveness lacks behind. This could be because their model is trained on only a subset of the training data where the comments are countering specific points in the post.

\paragraph{Analysis} 

As mentioned, conclusions may appear in arguments implicitly, which we expect to correlate with the quality of the generated counters: the more explicit the conclusion, the better the generated counters. Moreover, we hypothesize that, the longer an argument is, the more important the conclusion inference is in counter-argument generation. We empirically investigate these two hypotheses by comparing the performance of the counter-argument generation models subject to \emph{argument length} (in terms of the number of tokens) and to the degree of \emph{conclusion implicitness} (in terms of the maximum similarity between the ground-truth conclusion and premises). In particular, for both dimensions, we sorted the a sample of 2k instances from the test set accordingly and split it into five subsets of equal size. We then compare the BERT F$1$-score and contrastiveness score of {\em Joint One-seq} against {\em BART-based w/o Conclusion} and {\em BART-based w/ Conclusion} on the respective subset.

Figure~\ref{analysis-graphs} shows the scores for all three models at different levels of argument length and conclusion implicitness. In Figure~\ref{analysis-graphs}a, we see that the baseline's contrastiveness drops from 0.53 to 0.43 the longer the argument gets, while the scores for \emph{BART-based w/ Conclusion} fluctuate relatively around 0.57. In contrast, our approach achieves scores between 0.64 and 0.73, indicating the benefit of the explicit modeling of conclusions. Figure~\ref{analysis-graphs}c suggests that the more direct the conclusion is formulated in the premises, the better \emph{BART-based w/o Conclusion}'s contrastiveness score gets, and vice versa for \emph{BART-based w/ Conclusion} model. We observe an unexpected drop in scores for arguments where conclusions have an average similarity of 0.7 to the premises. Upon inspection, we found that the baselines tend to copy parts of the premises with slight rephrasing. Our approach, however, maintains high scores and also benefits from the clear formulation of the conclusion in premises since it helps to generate better conclusions. 

Lastly, looking at BERT F$1$-scores in Figures~\ref{analysis-graphs}b and \ref{analysis-graphs}d, we notice a drop across all approaches when arguments get longer. Similarly, the more apparent the conclusion in the premises, the better the scores get.

\subsection{Manual Evaluation}

To gain more reliable insights into the performance of our approaches, we designed a human evaluation study to measure the quality of the generated counters in terms of relevance to the input argument and the correctness of their stance. In a second study, we also let humans assess the validity of the generated conclusions.

\paragraph{Counter-Arguments} 

We selected 100 test set arguments randomly along with the counters generated by the two variations of our approach, \emph{Joint One-seq w/ Stance} and \emph{Joint Two-decoders w/ Stance}, as well as by two baselines, \emph{BART-based w/o Conclusion} and \emph{Pipeline-based}. Using the UpWork platform, we recruited three human annotators who are proficient in English with a job success of more than 90\%. We presented them the 100 arguments together with the texts of the four given counters, shuffled pseudo-randomly for each argument. For each argument, we then asked them to rank the texts based on their adequacy of being a counter-argument to the input argument, where we defined adequacy as follows: 
\begin{quote}
\em An adequate counter is a text that (1) carries an argumentative and coherent language and (2) clearly represents an opposing stance to one of the main points in the input argument. 
\end{quote}

\noindent
Additionally, the annotators should provide comments describing their decision regarding the counters ranked first (the best) and fourth (the worst). Computing the inter-annotator agreement using Kendall's $\tau$ results in an average of 0.32 (ranging from 0.32 to 0.43), while we observed majority agreement on full ranks between the annotators in 78\% of the evaluated cases.


\begin{table}[t!]%
	\centering%
	\small
	\renewcommand{\arraystretch}{1}
	\setlength{\tabcolsep}{3pt}%
	\begin{tabular}{lrr}
		\toprule
        		\bf Counter Generation Approach & \bf Average $\downarrow$ & \bf Majority  $\downarrow$  \\
		\midrule
		{BART-based w/o Conclusion}  & 2.56 & 2.54 \\
		{Pipeline-based} w/ Stance				& \bf 2.38 & 2.31 \\
		\addlinespace
		{Joint-based One-seq} w/ Stance		& 2.39 & \bf 2.26 \\
		{Joint-based Two-decoders} w/ Stance	& 2.65 & 2.72\\
		\bottomrule
	\end{tabular}%
	\caption{Manual evaluation: The \emph{average} and \emph{majority} rank of the counters generated by our approaches and baselines.}
	\label{table-counter-manual-eval}%
\end{table}

\begin{figure}
	\centering
	\includegraphics{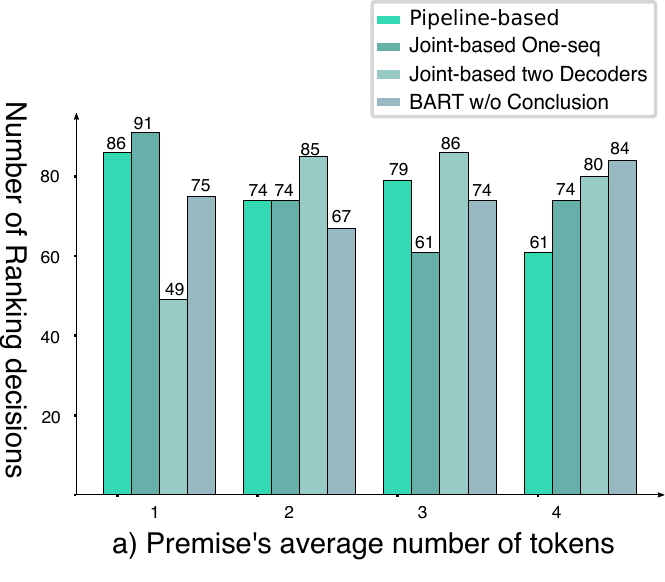}
	\caption{A histogram of Ranking scores that each of the manually evaluated approaches achieved.}
	\label{learn-to-use-labels-yourself}
\end{figure}

Table \ref{table-counter-manual-eval} reports the mean of the average and majority ranks achieved by each approach. When considering cases with majority agreement, our model {\em Joint One-seq w/ Stance} performs best (mean rank 2.26). This also can be seen in Figure~\ref{learn-to-use-labels-yourself}, where we plot the rank distribution for all approaches. In 55\% of the cases, the approach generated counters that were ranked either first or second. However, the variation with two decoders falls short compared to all others (mean rank 2.72). This suggests that sharing only the encoder between the two tasks is not enough to generate relevant counters. Also, as indicated before, not being able to prompt the generated conclusions limits the diversity of candidates in the stance-based ranking component. Finally, we see that the \emph{pipeline-based} baseline equipped with our ranking component is almost on par with our approaches, indicating the importance of promoting stance correctness.

\begin{table}[t!]%
	\centering%
	\small
	\renewcommand{\arraystretch}{1}
	\setlength{\tabcolsep}{8pt}%
	\begin{tabular}{lr}
		\toprule
		  \bf Conclusion Generation Approach & \bf Validity $\uparrow$ \\
		\midrule
		 {Pipeline-based} w/ Stance				& 1.42 \\
		{Joint-based One-seq} w/ Stance	& 1.91  \\
		 {Joint-based Two-decoders} w/ Stance	& 2.03 \\
		\midrule
		Ground-truth Conclusions   & \bf 2.39  \\
		\bottomrule
	\end{tabular}%
	\caption{Manual evaluation: Average \emph{validity} score from 1 (non-valid) to 3 (valid) of the conclusions generated by our approaches and the baseline compared to the ground truth.}
	\label{table-conclusion-manual-eval}%
\end{table}

\paragraph{Conclusions} 

To investigate whether the joint learning of conclusion and counter-argument generation leads to more valid conclusions, we designed another human evaluation study, for which we defined validity in a simple way:
\begin{quote}
\em A conclusion is valid if humans can infer it from the input argument. 
\end{quote}

\noindent
For 50 random arguments, we selected their ground-truth conclusion as well as two conclusions generated by the two variations of our approach and the best baseline (\emph{Pipeline-based w/ Stance}), summing up to seven conclusions per argument. We hired two annotators through UpWork again. We asked them to read each argumentsand to evaluate the validity of each conclusion on a 3-point Likert scale, where $3$ means that they strongly agree that the conclusion can be inferred and $1$ means they strongly disagree. The agreement of the two annotators was 0.46 in terms of Cohen's $\kappa$.

Table \ref{table-conclusion-manual-eval} shows the average scores achieved by each evaluated model. With 1.42, the {\em pipeline-based} approach is notably worse than the others, indicating the advantage of multitask learning for  conclusion and counter generation. The best score is achieved by {\em Joint Two-decoders w/ Stance} (2.03), being only 0.36 points below the ground-truth conclusion's score. Given the low effectiveness of this model on the counter-argument generation task, we assume that the training process optimized more towards generating conclusions, especially since the task may be easier than generating counters. A better weighting scheme for the two tasks may alleviate this in future work. 
\section{Conclusion}
In this paper, we studied the task of counter-argument generation, considering the role of the argument's conclusion. We argued that automatically generating counter-arguments becomes more challenging when the argument's conclusion is implicit, mandating explicit modeling. To validate our claims, we propose an approach that jointly learns to generate the conclusion and a counter for a given argument and compare it to baselines with no explicit conclusion modeling. We then realize our approach in two ways, both using transformer-based models. Moreover, our approach explicitly ensures that the generated counters have a correct stance through a stance-based ranking component. Our results show that, although far from perfect, the joint learning of the two tasks leads to better counters and more valid conclusions of the input argument when compared to baselines.

\section{Limitation}
In our experiments, we only experiment with the BART as the underlining transformer-based model. However, a stronger experiment setting could demonstrate the gain of modeling the conclusion across several transformer-based models, like GPT and T5. We didn't explore all possible weighting schemes for the two jointly learned tasks in our multitask setting. A potential direction could consider a more systematic evaluation of different schemes. Our models are limited by the quality of the data we use. To this end, we build on the assumption that CMV commentators rebut the original post's conclusion. However, this might not always be a valid assumption.

\section{Ethics}
Although our experiments demonstrate the role of conclusion in counter-argument generation, we believe this task is far from resolved. We are aware that issues such as faithful text generation must be considered when working with language models to avoid misinformation. We believe that a fact-checking component should accommodate any text generation model. The sole goal of our experiments is to highlight the potential of conclusion inference as part of the counter-argument generation pipeline.

\bibliography{eacl23-conclusion-based-counter-argument-generation-lit}

\bibliographystyle{acl_natbib}

\appendix

\section{Example Analysis}
\begin{table}[t!]%
	\centering%
	\small
	\renewcommand{\arraystretch}{1.0}
	\setlength{\tabcolsep}{2.5pt}%
	\begin{tabular}{p{0.985\columnwidth}}
		\toprule
		{\bf Conclusion (title)}: Artificial General Intelligence does not need to be conscious to be more intelligent or effective than humans \\
		\midrule
		{\bf Premises (post):} In my opinion AGI needs the algorithms that make us intelligent and by having these missing pieces they will be a generally intelligent entity. I hear arguments that an AGI needs to have a subjective experience like a human being in order to be generally intelligent and I find this theory to be lackluster if the laws of physics can be computed than so can the algorithms that make general intelligence. After all we are information processing systems however, i am trying to explore all avenues of thought, even play devil's advocate and not get caught in some sort of echo chamber or even worse a spirit of the age.	\\
		\midrule
		{\em BART-based w/o Conclusion:} \\ 
		{\bf Counter:} I don't think that AGI needs to have a subjective experience like a human being in order to be generally intelligent. I think it's more accurate to say that an AGI doesn't need to have subjective experience, but that doesn't mean that it can't have an objective experience.		\\
		\midrule
		{\em Pipeline-based w/ Stance:}\\
		{\bf Conclusion:} i need to be more intelligent in order to be an AGI.\\
		{\bf Counter:} intelligence is not determined by a subjective experience like a human being. it is determined by the laws of physics. an agi does not need to have an objective experience to be intelligent, but it does need to be able to understand the world around it.	\\
		\midrule
		{\em Joint-based One-seq w/ Stance:}\\
		{\bf Conclusion:} Scientific law is the only thing that can make AGI generally intelligent.\\
		{\bf Counter:} The problem with AGI is that we don't really know what it is that makes us intelligent. we have no idea how it works, what it's like to be an AGI, how it's different from a human being, or how it will work in the real world.\\
		\midrule
	\end{tabular} 
	\caption{An example argument (conclusion + premises) taken from {\em CMV} showing how the conclusion is implicitly mentioned in the body}
	\label{table-full-examples}
\end{table}

\paragraph{Qualitative Analysis}

Table \ref{table-full-examples} shows an example argument discussing {\em Artificial Intelligence} along with counters generated by the two baselines as well as by our approach {\em Joint-based One-seq w/ Stance}. {\em BART-based w/o Conclusion} rephrases sentences from the input argument without generating a proper counter, possibly due to the ignorance of the conclusion. While the {\em pipeline-based} baseline equipped with our ranking component generates a somehow relevant conclusion, its counter still vague and doesn't clearly oppose the argument's stance. Finally, {\em Joint-based One-seq} infers a conclusion that addresses the main point of the input argument ({\em Scientific law}), and counter it by pointing out the difficulty of defining {\em intelligent }, making it hard to be measured.

Upon exploring annotators' comments that justified their decisions of what is the best/worse counter, we identified some patterns. For example, {\em Joint-based One-seq} was most appreciated, because it generated argumentative and coherent counters that sometimes offered new perspectives. In contrast, the cases in which the model's output was ranked worst happen mainly due to being vague, incoherent, or diverging from the main topic. The counters of {\em BART-based w/o Conclusion} were ranked worse due to coherences sometimes, but often due to not opposing to the input argument.

\section{Computing Infrastructure}
All our experiments are run inside an \texttt{ubuntu20.04} system using \texttt{Python 3.8.10}. The CUDA version is 11.2. We used one A100-SXM4-40GB GPU to train our models. The following libraries are required to run our experiments:

\begin{itemize}
	\item torch==1.11.0+cu113
	\item transformers==4.18.0
	\item flair==0.11
	\item spacy==3.3.1
	\item debater-python-api==3.5.8
\end{itemize}

\end{document}